\documentclass{article}

\usepackage{arxiv}

\usepackage[utf8]{inputenc} 
\usepackage[T1]{fontenc}    
\usepackage{hyperref}       
\usepackage{url}            
\usepackage{booktabs}       
\usepackage{amsfonts}       
\usepackage{nicefrac}       
\usepackage{microtype}      
\usepackage{lipsum}		
\usepackage{graphicx}
\usepackage[square,numbers]{natbib}

\usepackage{doi}

\usepackage{framed,multirow}

\usepackage{amssymb}
\usepackage{latexsym}

\usepackage{url}
\usepackage{xcolor}
\usepackage{booktabs}
\usepackage{amsmath}
\definecolor{newcolor}{rgb}{.8,.349,.1}

\usepackage{graphicx}
\usepackage{amsmath}
\usepackage{amssymb}
\usepackage{booktabs}
\usepackage{makecell}

\usepackage{amssymb}
\usepackage{pifont}
\newcommand{\cmark}{\ding{51}}%
\newcommand{\xmark}{\ding{55}}%
\newcommand{\dataset}{FACEMORPHIC}
\newcommand{\fulldataset}{FACEMORPHIC - Neuromorphic Face Dataset}

\newcommand{\revision}[1]{{\color{black}{#1}}}

\usepackage{xcolor}

\title{Spatio-temporal Transformers for Action Unit Classification with Event Cameras}

\author{ Luca Cultrera\\
	University of Florence\\
	\texttt{luca.cultrera@unifi.it} \\
	\And
	Federico Becattini\\
	University of Siena\\
	\texttt{federico.becattini@unisi.it} \\
	\And
	Lorenzo Berlincioni\\
	University of Florence\\
	\texttt{lorenzo.berlincioni@unifi.it} \\
	\And
	Claudio Ferrari\\
	University of Parma\\
	\texttt{claudio.ferrari2@unipr.it} \\
	\And
	Alberto Del Bimbo\\
	University of Florence\\
	\texttt{alberto.delbimbo@unifi.it} \\
}

\renewcommand{\shorttitle}{\textit{arXiv} Template}

\hypersetup{
	pdftitle={A template for the arxiv style},
	pdfsubject={q-bio.NC, q-bio.QM},
	pdfauthor={David S.~Hippocampus, Elias D.~Striatum},
	pdfkeywords={First keyword, Second keyword, More},
}

\begin{document}
	\maketitle
	
	\begin{abstract}
		As one of the most important applications in computer vision, face analysis has been studied from different angles in order to infer emotion, poses, shapes, and landmarks. Traditionally the research has employed classical RGB cameras to collect and publish the relevant annotated data. For more fine-grained tasks however standard sensors might not be up to the task, due to their latency, making it impossible to record and detect micro-movements that carry a highly informative signal, which is necessary for inferring the true emotions of a subject. Event cameras have been increasingly gaining interest as a possible solution to this and similar high-frame rate tasks. In this paper we propose a novel spatio-temporal Vision Transformer model that uses Shifted Patch Tokenization (SPT) and Locality Self-Attention (LSA) to enhance the accuracy of Action Unit classification from event streams.  We also address the lack of labeled event data in the literature, which can be considered one of the main causes of an existing gap between the maturity of RGB and neuromorphic vision models. In fact, gathering data is harder in the event domain since it cannot be crawled from the web and labeling frames should take into account event aggregation rates and the fact that static parts might not be visible in certain frames. To this end, we present FACEMORPHIC, a temporally synchronized multimodal face dataset composed of both RGB videos and event streams. The dataset is annotated at a video level with facial Action Units and also contains streams collected with a variety of possible applications, ranging from 3D shape estimation to lip-reading. We then show how temporal synchronization can allow effective neuromorphic face analysis without the need to manually annotate videos: we instead leverage cross-modal supervision bridging the domain gap by representing face shapes in a 3D space.
		Our proposed model outperforms baseline methods by effectively capturing spatial and temporal information, crucial for recognizing subtle facial micro-expressions.
	\end{abstract}


	\begin{figure}[t]
		\centering
		\includegraphics[width=.7\linewidth]{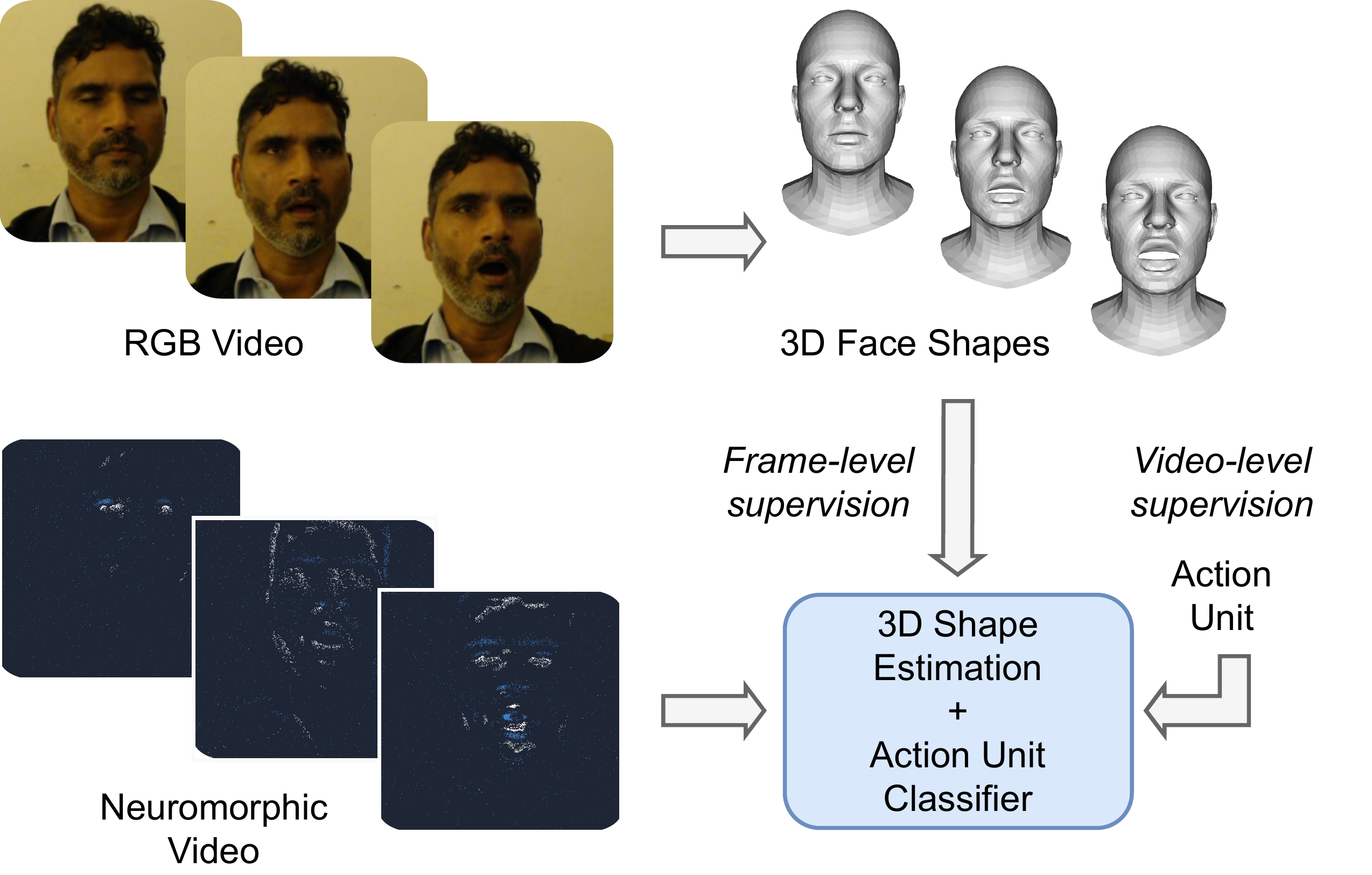}
		\caption{We leverage cross-modal supervision obtainable from temporally synchronized RGB and Event streams to analyze faces using neuromorphic data. By extracting 3D face shape coefficients with standard RGB vision models, we can improve the training of event-based models without additional manual labeling.}
		\label{fig:eyecatcher}
	\end{figure}

	\section{Introduction}
	\label{sec:intro}
	
	
	\revision{The interpretation of human expressions is a key feature of several computer vision applications. Tasks such as emotion recognition, 3D face modeling, or identity recognition all rely on face analysis, as confirmed by the large amount of research involving it. Several open-source models for face analysis applications are now provided as off-the-shelf tools for face detection \citep{dlib09}, landmark detection \citep{bulat2017far} and gaze estimation \citep{zhang2017mpiigaze}, just to name a few.}
	
	\revision{
		Still, achieving a detailed understanding of facial expressions is inherently challenging due to the continuous micro-movements generated by facial muscles, which can occur rapidly and suddenly. The activations of facial muscles, typically referred to as Action Units, have been intensively studied for their association with the underlying emotions. This research has led to the development of the Facial Action Coding System (FACS) \citep{ekman1978facial}, which maps Action Units to specific emotions. The challenge lies in the fact that micro-expressions and Action Units can last as little as 80ms \citep{yan2013fast}, making it difficult to fully capture their evolution using standard RGB cameras, which operate at 25/30 FPS.}
	
	\revision{In order to effectively model these facial dynamics some previous works \citep{zhao2023more} have been employing high-framerate cameras with the downside of producing orders of magnitude more frames that then need to be processed.
		In this paper we follow a different approach based on neuromorphic vision.} Neuromorphic sensors, often referred to as event cameras, are biologically inspired sensors that produce asynchronous streams of events rather than synchronous streams of frames. An event is defined as a local change in illumination at a pixel level and can be fired asynchronously at a microsecond rate.
	
	
	\revision{Initially event cameras gained attention, especially in the field of robotics, thanks to their low power consumption, extremely low latency, high dynamic range, and absence of motion blur. Recently, outside the robotics field, a few works employed these sensors for face analysis \citep{becattini2024neuromorphic, becattini2022understanding, berlincioni2023neuromorphic, berlincioni2024neuromorphic, ryan2023real, shariff2023neuromorphic, bissarinova2023faces}, often focusing on Industry 4.0 applications such as drowsiness estimation \citep{shariff2023neuromorphic} or facial reaction analysis \citep{becattini2022understanding}.}
	
	\revision{However, using neuromorphic cameras for facial analysis comes with its limitations. Decades of research on RGB images and videos cannot be fully applied to the event-based domain. Models trained on RGB data are ineffective on neuromorphic streams due to the significant differences between the data types \citep{becattini2024neuromorphic}. Even when events are aggregated into frames \citep{mueggler2017fast, innocenti2021temporal, nguyen2019real}, the substantial domain shift limits the performance of these models, often requiring new architectures to be developed for tasks that are nearly solved in the RGB domain, such as face detection. This underscores the importance of annotated neuromorphic datasets, although only a few, limited ones are currently publicly available \citep{bissarinova2023faces, berlincioni2023neuromorphic}. }
	
	\revision{Several published works attempt to solve this data scarcity issue by using simulators \citep{rebecq2018esim, hu2021v2e} in order to convert RGB videos into event streams. By converting data to the event domain existing large RGB datasets can be leveraged to train novel event-based models.}
	\revision{While some recent works have adopted this approach for face detection \citep{becattini2022understanding, berlincioni2023neuromorphic}, the conversion process is slow, and the resulting streams can contain spurious events caused by compression artifacts \citep{berlincioni2024neuromorphic}. Additionally, fast movements like micro-expressions, when captured at low frame rates in video datasets, may not translate properly into event streams, as the necessary signal is missing from the original data.}
	
	\revision{To address this challenge of obtaining high quality annotated data we introduce \dataset{}, a new multimodal dataset featuring subjects performing Action Units, captured with temporally synchronized RGB and event cameras. Through this temporal alignment, we show that it is possible to generate supervision signals directly from the RGB stream by representing the face in a camera-invariant reference frame. This is achieved by characterizing the face with the coefficients of a 3D Morphable Model, fitted in the RGB domain.}
	We therefore leverage two distinct sources of supervision: video-level supervision, obtained through manual labeling while collecting the dataset (each video contains a different action unit); and a frame-level cross-domain supervision, derived by applying traditional computer vision models on the RGB frames.
	In this sense, the proposed cross-domain supervision resembles the idea of distillation \citep{hinton2015distilling}, where typically a large, well-trained network (\textit{teacher}) is used to supervise the training of a smaller network (\textit{student}).
	We show that cross-domain supervision used for regressing 3D face deformation coefficients frame per frame, aids the learning process of an Action Unit classifier. The idea is shown in Fig.~\ref{fig:eyecatcher}.

	\revision{A preliminary study our collected dataset has been recently carried out in \citep{becattini2024neuromorphic}. However, we only tested simple baseline methods to prove the effectiveness of modeling facial action units with an event camera and of using the cross-modal supervision. In this paper we expand our study, by introducing a novel model specifically tailored to take into account spatio-temporal cues from event data. The proposed approach is a two-stage Vision Transfomer model, composed of a spatial transformer module processing individual frames and a temporal transformer module designed to capture temporal dynamics across video frames. The spatial transformer module takes inspiration by \citep{lee2021vision} by including Shifted Patch Tokenization (SPT) and Locality Self-Attention (LSA) to enrich the semantics of the visual tokens and improve the spatial attention mechanism. The proposed approach is still compatible with our cross-modal supervision, being it model-agnostic.}
	\revision{In addition, we also investigate whether the learned model can act as a form of pre-training for other event-based face analysis tasks. In particular, we report noticeable improvements in the downstream task of emotion classification. This suggests that learning to generate 3D deformation coefficients precisely describing the face yields generic features that can be easily adapted to other face-related tasks in the event domain. We believe this to be particularly important, considering the scarce availability of event-based face datasets.}
	
	In summary, the main contributions of our paper are:
	\begin{itemize}
		
		\item \revision{We propose a novel Vision Transformer model combining spatial and temporal information, specifically designed to enhance the performance of vision transformers in the detection of fine-grained facial Action Units from event-based data.}
		
		\item \revision{To demonstrate the capabilities of our model, We present \fulldataset{}, the first multimodal dataset for action unit detection with temporally synchronized neuromorphic and RGB videos. We collected more than 4 hours of recordings for each modality, including Action Units performed by 64 users.}
		
		\item \revision{In addition to the proposed model, we introduce a learning framework for paired RGB and event data, involving both video-level and cross-domain supervision, applied frame-by-frame thanks to information derived from RGB vision modules providing 3D information.}

		
		\item \revision{We demonstrate that the proposed model, trained on our \fulldataset{} dataset, can act as a form of pre-training for event-based downstream tasks: by using action unit classification as a proxy task, we achieve state of the art results on the NEFER dataset emotion recognition benchmark.}
		
	\end{itemize}

	\section{Related Works}
	
	\paragraph{\textbf{Neuromorphic Vision}}
	\revision{Neuromorphic vision refers to a class of sensors and data acquisition techniques that utilize event cameras, designed based on an innovative, biologically-inspired vision model \citep{delbruckl2016neuro, posch2014retino}. Unlike conventional vision systems, which generate synchronized frame sequences at fixed intervals, this approach produces an asynchronous stream of events.}
	\revision{An \textit{event} is the occurrence of a local change in brightness that is large enough to be detected by a sensor operating at an extremely high temporal resolution (in the order of microseconds) with very low latency \citep{lichtsteiner2008asynch}. A key advantage of these neuromorphic sensors is their ability to remain inactive unless a localized brightness change occurs, which conserves resources and lowers bandwidth usage \citep{gallego2020event}.}
	
	\revision{As the technology behind these sensors improves and as consequentially they get more affordable, in several fields,
		such as robotics \citep{mueggler2017event,9813406}, tracking \citep{seok2020robbust,renner2020event, iddrisu2024evaluating}, lip-reading \citep{bulzomi2023end, savran2018energy}, and object detection \citep{DBLP:journals/tim/LiuXYYY23,8593805}, the benefits of event cameras can be fully appreciated \citep{gallego2020event,delbruckl2016neuro,ramesh2019dart}.}
	
	\revision{In addition to the previously mentioned fields, the distinctive characteristics of event cameras are well-suited for analyzing facial expressions at high temporal resolutions.
		\cite{iddrisu2024evaluating} demonstrated the effectiveness of integrating event data into conventional Convolutional Neural Network models for face and eye detection tasks, showing strong performance across multiple event camera datasets.	
		However, in \citep{berlincioni2024neuromorphic}, human emotions are evaluated based on valence and arousal by utilizing an event-data simulator to transform RGB videos. While this synthetic approach provides insights, it does not fully exploit the potential of event cameras. Despite growing interest in the subject, only a few datasets capturing facial dynamics using real event cameras are available in the literature \citep{bissarinova2023faces, savran2020face, moreira2022neuromorphic, lenz2020event, becattini2022understanding, berlincioni2023neuromorphic, chen2020eddd}.}
	The authors of~\citep{savran2020face} focus on the task of face pose alignment providing a dataset of 108 videos of several head rotations with varying intensities for a total of 10 minutes of footage.
	In \citep{lenz2020event} instead, the presented event-data is collected for eye blink detection. It consists of 48 videos for a cumulative duration of about 13 minutes.
	\cite{becattini2022understanding} present a dataset of 455 videos of facial reactions where the recorded users react to garment images
	Finally in \citep{berlincioni2023neuromorphic} the authors collect a dataset composed of paired \textit{RGB - event data} for emotion recognition providing also facial bounding box and landmark annotations in addition to emotion labels.
	
	In this paper we propose a new dataset, \dataset{}, to effectively address face analysis by action unit classification. Comprising more than 4 hours of videos, it is the largest existing dataset related to human facial expressions and emotions, as the datasets collected in  \citep{berlincioni2023neuromorphic} and \citep{becattini2022understanding} have a total extent of 13 and 75 minutes. At the same time, we provide labels covering a set of 24 action units, instead of categorizing videos in binary reaction \citep{becattini2022understanding} or 7 basic emotions \citep{berlincioni2023neuromorphic}.
	
	
	
	\paragraph{\textbf{Facial Action Units}} The Facial Action Coding System \citep{ekman1978facial} refers to a set of popular facial behavior signs judgment methods. It is an exhaustive anatomical-based system that encodes various facial movements by the combination of basic Action Units (AU). This set of AUs constitutes a sort of alphabet for the, more complex, human face expressions.
	Action Units define certain facial configurations caused by the contraction of one or more facial muscles, and they are independent of the interpretation of emotions.
	In the human-computer interaction field the use of this taxonomy enabled a large range of applications such as in security settings \citep{Salah2010CommunicationAA,1678017}, commercial contexts to estimate consumer reaction to products \citep{zhi2020comprehensive,7374704, becattini2021plm} \revision{and clinical studies for pain detection and patient monitoring \citep{LUCEY2012197,Littlewort2007FacesOP, sugiyanto2024depression, parikh2024exploring}. Particularly, \citep{sugiyanto2024depression} proposed a CNN-Poolingless framework using 14 AU intensity features, showing improved depression classification. Similarly \citep{parikh2024exploring} emphasized the use of AUs as biomarkers for depression, using time-series models and Principal Component Analysis (PCA) to analyze expression patterns linked to sadness and happiness, demonstrating the potential of facial analysis for accurate depression diagnosis.
		
		\cite{enabor2024action} proposed a Weakly Supervised Multi-Label approach with Large Loss Rejection (LLR) to handle noisy labels, achieving promising results in AU recognition. In the realm of micro-expression analysis, where expressions are brief and subtle, \cite{yang2024micro} developed a dual-path transformer network that successfully extracted key features from apex frames, enhancing classification performance. Finally, another method introduced a self-attention spatiotemporal fusion model (SAtt-STPN) to capture both intra- and inter-frame AU relationships, outperforming state-of-the-art models in AU detection \citep{liang2024facial}.
	}
	
	\paragraph{\textbf{3D Morphable Models}} Since the seminal work of Blanz and Vetter~\citep{blanz2023morphable}, the 3D Morphable Model (3DMM) has been extensively used in the field of face analysis to address a variety of tasks. The 3DMM is a statistical model of shapes, and is usually learned from a set of 3D faces in dense correspondence. Its expressive power depends mostly on the training data, in which direction efforts have been made to build large-scale datasets~\citep{booth20163d} or augmenting the spectrum of possible deformations~\citep{principi2023florence}. Several variants have been proposed for learning a 3DMM, ranging from the standard PCA model~\citep{paysan20093d}, to other solutions based for example on Gaussian process~\citep{luthi2017gaussian}, Dictionary Learning~\citep{ferrari2017dictionary}, multilinear wavelets~\citep{brunton2014multilinear} and so on. Lately, deep networks have been employed as well to learn 3DMMs thanks to their highly non-linear and powerful generation quality~\citep{bouritsas2019neural,ranjan2018generating,tran2018nonlinear,zheng2022imface}. It has been widely used for face reconstruction~\citep{galteri2019deep,gecer2019ganfit,tuan2017regressing,wu2019mvf} or recognition~\citep{hu2016face,an2018deep,koppen2018gaussian,ferrari2015dictionary}, expression and Action Units recognition or generation~\citep{shi2020pose,chang2018expnet,otberdout2023generating,ariano2021action,wang2021improved}. Overall, the related literature reveals the 3DMM still as a state-of-the-art technique for face modeling with several possible applications.

	\section{Motivation}
	Event cameras can capture illumination changes, and thus motion, at an extremely fast pace. Being able to effectively analyze faces at such a data rate would allow us to precisely characterize expressions and their underlying emotions.
	However, face analysis in the neuromorphic domain is proceeding slowly despite increasing interest \citep{becattini2024neuromorphic}, motivated by a few preliminary results indicating its usefulness in industrial scenarios \citep{ryan2023real, shariff2023neuromorphic} and its effectiveness over RGB \citep{becattini2022understanding, berlincioni2023neuromorphic}.
	Driven by the desire to realize event-based face analysis models, we collected \dataset{} to foster research on this topic. The dataset we collected is temporally synchronized across modalities, meaning that it is possible to obtain different representations of the same face, captured from two sensors - the RGB and the neuromorphic one. As a consequence, any available temporal annotation can be transferred from one media to another.
	
	The same cannot be said for spatial annotations, such as bounding boxes or facial landmarks, since the sensors are uncalibrated and the subject can sit freely in front of the cameras. Working under the assumption of uncalibrated cameras though opens up the problem of precisely annotating data. Spatially labeling faces in RGB frames is almost trivial: off-the-shelf face detectors and landmark detectors can nowadays be effectively used to gather labels automatically, without any manual labor. Conversely, such vision models cannot be directly applied to event frames.
	\cite{becattini2024neuromorphic}, have recently shown that the outputs of RGB models, such as face and landmark detectors, are likely to fail on neuromorphic data, pointing out that when a small amount of motion is present, no meaningful output is obtainable, whereas when faces are sufficiently visible in the event frame, some low-confidence and imprecise detection might be still be gathered. In both cases, such models are unlikely to be of any practical use.
	
	
	In virtue of these considerations, our goal is to leverage temporally synchronized modalities to transfer automatically generated labels from RGB frames onto events, exploiting a characterization of the face that goes beyond a spatial reference system grounded on either the RGB or the event frame. To this end, we will infer the coefficients of a 3D Morphable Model for event-based frames, bypassing the generation of bounding boxes and facial landmarks, that are usually required to obtain such information.
	In this way, by lifting the annotations from the frame to a pose-agnostic 3D representation we can effectively describe the shape of the face and label event streams with a cross-modal supervision.
	

	\newcommand{\xmarktable}{-}
	
	\begin{table*}[t]
		\resizebox{\linewidth}{!}{
			\begin{tabular}{l|ccccccccc}
				Dataset & Year & Videos & Duration & Users  & Resolution & Task & Open Source & ~~~RGB~~~ & Synch. \\
				\hline
				\citep{savran2020face} & 2020 & 108 & 10 min. & 18 & 304$\times$204 & Face Pose Align. & \xmarktable & \xmarktable & \xmarktable  \\
				\citep{lenz2020event} & 2020 & 48 & 13 min. & 10 & 640$\times$480 & Face Detection & \xmarktable & \xmarktable & \xmarktable  \\
				\citep{berlincioni2023neuromorphic} & 2023 & 609 & 13 min. & 29 & 1280$\times$720 & Emotion Class. & \ding{51} & \ding{51} & \xmarktable \\
				\citep{savran2018energy} & 2018 & 360 & 28 min. & 18 & 304 $\times$ 240 & Voice Activity Det. & \xmarktable & \xmarktable & \xmarktable \\
				\citep{becattini2022understanding} & 2022 & 455 & 75 min. & 25 & 640$\times$480 & Reaction Class. & \xmarktable  & \ding{51} & \xmarktable \\

				\citep{chen2020eddd}  & 2020 & 260 & 86 min. & 26 & 346$\times$260 & Driving Monitoring & \ding{51} & \ding{51} & \xmarktable \\
				\citep{moreira2022neuromorphic} & 2022 & 432 & 180 min. & 40 & - & Identity Recognition & \xmarktable & \xmarktable & \xmarktable \\
				
				\citep{tan2022multi} & 2022 & 200 & 231 min. & 40 & 346$\times$260 & Lip Reading & \ding{51} & \ding{51} & \xmarktable \\
				\citep{bissarinova2023faces}~ & 2023 & 3889 & 689 min. & 73 & 408$\times$360 & Face Detection & \ding{51} & \xmarktable & \xmarktable \\
				\hline
				FACEMORPHIC & 2024 & 3148 & 248 min. & 64 & 1280$\times$720 & Action Unit Class. & \ding{51} & \ding{51} & \ding{51}
			\end{tabular}
		}
		\caption{Comparison with other neuromorphic face datasets.} 
		
		\label{tab:datasets}
	\end{table*}
	
	\section{The \dataset{} Dataset}
	In this section, we present \fulldataset{}, that we collected for our experiments. To the best of our knowledge it is the first multimodal RGB and Event dataset for Facial Action Unit classification. All the videos in the dataset are temporally synchronized across modalities and are recorded with a commercial USB RGB camera and a Prophesee Evaluation Kit 4 (EVK4), equipped with the IMX646 neuromorphic sensor. The RGB camera and the event camera have different resolutions, respectively of $640 \times 480$ and $1280 \times 720$.
	Each data acquisition session was performed by recording with the two cameras the following for each user: (i) an initial recording, where the user can freely interact with the environment, speak and look around; (ii) separate recordings of the user performing 24 different Action Units - each recording is repeated twice; (iii) four recordings of the subject reading a short sentence drawn at random.
	Overall, we collected a total of 3148 videos, corresponding to 4.13 hours of recording for each modality. In the dataset, 64 users are present (16 females and 48 males), ranging from age 18 to 67. 
	
	The 24 Action Units recorded in the dataset include 18 micro-actions related to face muscle activations plus 6 macro-actions involving head movements. All the Action Units have been selected among the Facial Action Coding System (FACS) \citep{ekman1978facial} and were chosen in order to include the Action Units that are usually studied in the vision literature.
	In particular, we include all the Action Units used in the popular DISFA\footnote{Action Units 1, 2, 4, 6, 9, 12, 25, 26} \citep{mavadati2013disfa} and BP4D\footnote{Action Units 1, 2, 4, 6, 7, 10, 12, 14, 15, 17, 23, 25} \citep{zhang2014bp4d} datasets. In addition we included also the ones related to eye movements (AU 43 and 45) and head movements (AU 51, 52, 53, 54, 55, 56).
	The complete list of Action Units present in \dataset{} is: 1, 2, 4, 6, 7, 9, 10, 12, 14, 15, 17, 23, 24, 25, 26, 27, 43, 45, 51, 52, 53, 54, 55, 56.
	In Tab. \ref{tab:datasets} we present a comparison between \dataset{} and existing facial neuromorphic datasets from the literature.
	Differently from all the other datasets, \dataset{} is the only dataset providing synchronized RGB+Event data. This enables cross-modal supervision, thus allowing us to learn complex facial dynamics without costly annotation procedures.
	It must be noted that most datasets are either extremely small or are recorded with low-resolution sensors. The only existing large-scale dataset is FES \citep{bissarinova2023faces}, which nonetheless addresses only face detection at a low resolution and does not come with RGB data.
	\dataset{} is going to be publicly released for research purposes. The release will include also facial bounding boxes and 3D landmarks estimated from the RGB frames using Face Alignment \citep{bulat2017far} and the face shape coefficients for the 3D Morphable Model fitted on the landmarks, as described in Sec. \ref{sec:3dmm}. We are also releasing the sentences read by the users in the data acquisition process. Making subjects read sentences was intended as a way to let users move their faces in a natural way, but since some interest in lip reading with event cameras has been shown in the literature \citep{tan2022multi, bulzomi2023end}, we release the annotations and we leave this to further investigation in future work. In all our experiments we defined an 80-20 split between train and test videos.
	
	\paragraph{\textbf{Dataset Contribution and Responsibility to Human Subjects}}
	\revision{The collection of the FACEMORPHIC dataset has been carried out by informing all the subjects of the purpose of the dataset and that it will be distributed for research purposes. We have asked all participants to sign an agreement compliant with European privacy guidelines.
		We plan to make the dataset publicly available for research use. The dataset will be provided upon request and third party users will be requested to sign a responsibility agreement, upon which an individual is going to be designated as responsible for handling any further request concerning the dataset. This is necessary since subjects recorded in the dataset have the right to request the removal of the recordings from the dataset at any time. The access to the data will therefore be granted to research groups that designate an individual who agrees to apply any requested change in the future.}

	\begin{figure*}[!t]
		\centering
		\includegraphics[width=0.7\linewidth]{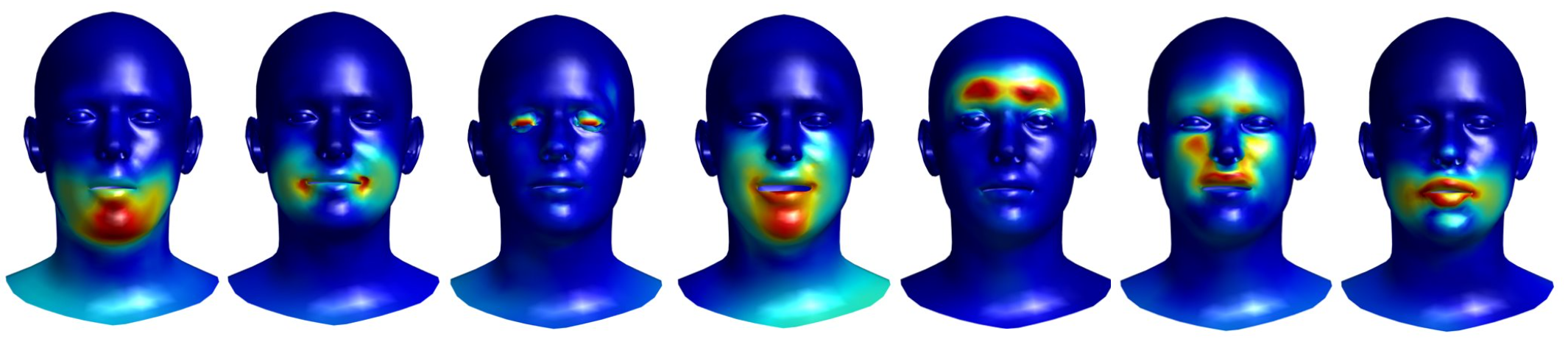}
		\caption{Example of AU-3DMM components learned from the D3DFACS dataset. The heatmaps show the spatial extent of the deformation (\textcolor{red}{red}=high, \revision{blue}=no deformation). The learned components capture AU specific facial movements.}
		\label{fig:AU-3DMM}
	\end{figure*}

	\section{Leveraging 3D Faces for Cross-Modal Labeling}
	In order to effectively train an Action Unit classifier with event data, we rely on video-level supervision (e.g. the AU class label) as well as a cross-modal supervision at frame level. Such supervision comes from face shape coefficients estimated from the RGB with a 3D Morphable Model (3DMM). \revision{Being the 3D face shape encoded in the 3DMM parameters, we can directly transfer this supervision by associating them with event frames, given that modalities are temporally aligned.} In the following we first provide details about the 3D Morphable Model used to estimate face shapes and we then motivate the annotation transfer across modalities.
	
	\subsection{3D Morphable Model}
	\label{sec:3dmm}
	The 3DMM is a statistical deformation model for 3D faces, firstly proposed by Blanz and Vetter~\citep{blanz2023morphable}. A 3DMM is built by learning a low-dimensional space from a set of densely registered 3D faces. The learned basis vectors are used to parameterize the shape (and optionally texture) space and synthesize new faces as:
	
	\begin{equation}
		\mathbf{S} = \mathbf{T} + \mathbf{C}\mathbf{\alpha}
		\label{eq:3dmm}
	\end{equation}
	where $\mathbf{S}\in \mathbb{R}^{N \times 3}$ is a 3D face of $N$ vertices, $\mathbf{T}\in \mathbb{R}^{N \times 3}$ is a template 3D face, $\mathbf{C}\in \mathbb{R}^{3N \times K}$ are the shape bases and $\alpha \in \mathbb{R}^K$ are the deformation coefficients.
	
	Depending on the \revision{number of samples, variability and type of deformations included in the training set of 3D faces,} different bases of \textit{deformation components} can be learned. \revision{These define what kind of deformations can be applied by the learned model.} For example, the Basel Face Model~\citep{paysan20093d} is built from a set of 200 faces in neutral expression, and the learned deformations encode structural facial traits \textit{e.g.} thin/large head, feminine/masculine etc. Differently, other methods such as the DL-3DMM described in~\citep{ferrari2015dictionary} or the FLAME~\citep{li2017learning} model, are learned from both neutral and expressive faces. In this way, the 3DMM can also replicate expression related deformations \textit{e.g.} mouth opening, eyebrow raising.
	
	For the purpose of this work, we build two separate 3DMMs, one for encoding structural identity deformations, and the other specific for encoding action-units activations. \revision{The reason for this choice is the following: if the facial shape of specific individuals gets encoded in the 3DMM parameters, then it would be difficult to estimate which coefficients contribute either to the reconstruction of the face shape or the deformations determined by the AU activation. Thus, to get rid of this nuisance, we first want to obtain an estimate of the 3D facial shape defining the individual. We do so by first reconstructing an approximate face shape assuming the first frame portrays the individual in neutral expression. Then, we can use this identity-specific model to estimate the facial deformation resulting from the AU activation in the subsequent frames.} To this aim, \revision{we first need to build two separate 3DMMs. Again, the reason is that the bases of deformation components capture all the variations in the training data. Thus, we need to learn a shape-specific 3DMM using only neutral faces, and another one specific for AUs.} We used the VOCASET~\citep{VOCA2019} and D3DFACS~\citep{Cosker2011AFV} datasets: the former includes 3D sequences of 12 actors performing facial expressions, while the latter 3D sequences of 10 actors performing AU activations. All meshes share the same (FLAME~\citep{li2017learning}) topology. We build the identity model $\mathbf{C}_I \in \mathbb{R}^{3N \times 22}$ (ID-3DMM) from the 22 joined actors, using only samples in neutral expression to learn the PCA space. To build the AU model $\mathbf{C}_{AU} \in \mathbb{R}^{3N \times K}$ (AU-3DMM), we first compute AU-specific deformation offsets between expressive and neutral scans; \revision{this is done in order to remove the identity component, leaving only AU-induced motions.} Then, we learn the the deformation components by applying the DL-3DMM algorithm of~\citep{ferrari2015dictionary} on such offsets. \revision{We chose this specific model as it can be learned very quickly, at the same time providing an effective modeling of facial deformations.} Various number of components $K$ have been tested. Some examples of how the learned AU-3DMM components can capture AU activations are shown in Fig.~\ref{fig:AU-3DMM}.
	
	Given the above models, we can then fit the 3DMM to the RGB frames, so to obtain identity and AU-specific deformation coefficients, $\alpha_I$ and $\alpha_{AU}$.  
	
	\subsection{Two-step 3DMM Fitting}\label{subsec:3dmm_fitting}
	In order to recover AU-specific deformation coefficients, it is first necessary to remove structural shape information related to the identity. To this aim, we perform a two-step 3DMM fitting, similar to~\citep{ferrari2018rendering}. Given that we are not interested in recovering accurate a 3D face reconstruction but only in capturing facial movements, we apply a landmark-based fitting algorithm. We chose to exploit the method in~\citep{ferrari2015dictionary} as it is extremely fast (solved in closed-form) and sufficiently accurate in modeling facial deformations. 
	
	First, $68$ facial landmarks $\mathbf{l} \in \mathbb{R}^{68 \times 3}$ are detected from the RGB frames using the method in~\citep{bulat2017far}, which also provides an approximate $z$ coordinate for each landmark. A corresponding set of 3D landmarks $\mathbf{L}_T \in \mathbb{R}^{68 \times 3}$ is labeled on a 3D face template $\mathbf{T} \in \mathbb{R}^{5023 \times 3}$ in FLAME topology. Given the detected landmarks and the template landmarks, the fitting is performed by first estimating an orthographic camera model from the landmark correspondence as
	$ \mathbf{A} = \mathbf{l} \cdot \mathbf{L}_T^\dagger$,
	%
	where $\mathbf{A} \in \mathbb{R}^{2 \times 3}$ is the camera matrix that contains 3D rotation, scale and shear parameters, and $\mathbf{L}_T^\dagger$ indicates the pseudo-inverse matrix. Then, we estimate the 2D translation $\mathbf{t} \in \mathbb{R}^{68 \times 2}$ as $\mathbf{t} = \mathbf{l} - \mathbf{A} \cdot \mathbf{L}_T$. Finally, the deformation coefficients $\alpha$ are estimated by minimizing the projection error between the detected landmarks and the back-projected template landmarks. This problem is cast as a regularized ridge-regression problem:
	
	\begin{equation}
		\min_{\alpha} \left \| \mathbf{l} - \mathbf{C}(\mathbf{A} \cdot \mathbf{L}_T + \mathbf{t} )\alpha \right \|_2^2 + \lambda \left \| \alpha \right \|_2
		\label{eq:fitting}
	\end{equation}
	which has a closed form solution as shown in~\citep{ferrari2015dictionary}. The parameter $\lambda$ controls the intensity of the deformation and serves to avoid excessive deformations of the template. A new 3D face is then synthesized using Eq.~\eqref{eq:3dmm}. 
	
	\paragraph{\textbf{Identity Model Fitting}}\label{subsubsec:id_fit}
	Assuming the first frame of each recorded video portrays the subject in neutral expression, we use the ID-3DMM to reconstruct and identity-specific 3D face on this frame. This is simply done by using the components $\mathbf{C}_I$ in Eq.~\eqref{eq:fitting}. As a result, we estimate identity-specific coefficients $\alpha_I$, so the new shape can be obtained as $\mathbf{S}_I = \mathbf{T} + \mathbf{C}_I\alpha_I$.
	
	\paragraph{\textbf{Estimating AU-specific deformation coefficients}}
	We use the estimated identity shape $\mathbf{S}_I$ to fit the AU model $\mathbf{C}_{AU}$ on all the subsequent frames of the RGB video. This strategy is intended to explicitly disentangle structural and AU-related shape deformations. Ideally, given that $\mathbf{S}_I$ captures the subject identity traits, if we use this 3D shape to fit the frames where the subject performs AU activations, the corresponding deformation coefficients $\alpha_{AU}$ should only capture the Action Units. To this aim, we repeat the process of Sec.~\ref{subsec:3dmm_fitting} yet this time using $\mathbf{S}_I$ in place of the template $\mathbf{T}$, and the AU-3DMM $\mathbf{C}_{AU}$ in place of $\mathbf{C}_I$. Hence, we collect a set of AU-specific coefficients $\alpha_{AU}$ for each frame.

	
	\subsection{Cross-Modal Labeling}
	Once face deformations, represented by the coefficients $\alpha_{AU}$, have been obtained, we can map them to event data without further manual annotation.
	In this paper, we simplify data transfer across modalities by generating event frames from raw events using an accumulation of 33ms. This yields event videos at 30 FPS, i.e. the same frame-rate of the RGB videos. Associating the coefficients to event frames thus is trivial, as we obtain the same number of frames in both modalities.
	
	Two important matters have to be taken into account. 
	First, since we strive to model micro-movements as fast as Action Units, a finer frame-rate could be desirable. In this case, frame association can be done by searching for the frame with the nearest timestamp. The annotation will not be dense, meaning that only 30 frames in each second will be annotated. The remaining frames will either be left without direct supervision or can be labeled by interpolating the temporally adjacent coefficients. We leave this investigation for future research.
	Second, we argue that adopting a frame-rate of 30 FPS does not affect the information collected by the event camera. In fact, if RGB cameras create frames by making a snapshot of the current intensity values for each pixel, event frames accumulate all the temporal information within the last $\Delta t=33ms$. This has the advantage of not increasing the frame number compared to RGB (hence, we have no increase in the amount of computation) while still being able to capture motions that happen at timesteps that are not multiples of the frame rate. 
	To aggregate events, we use the Periodic Frame Generation Algorithm implemented in the Prophesee SDK\footnote{https://docs.prophesee.ai/stable/concepts.html\#generating-frames-from-cd-events}.
	

	\begin{figure*}[t]
		\centering
		\includegraphics[width =\linewidth]{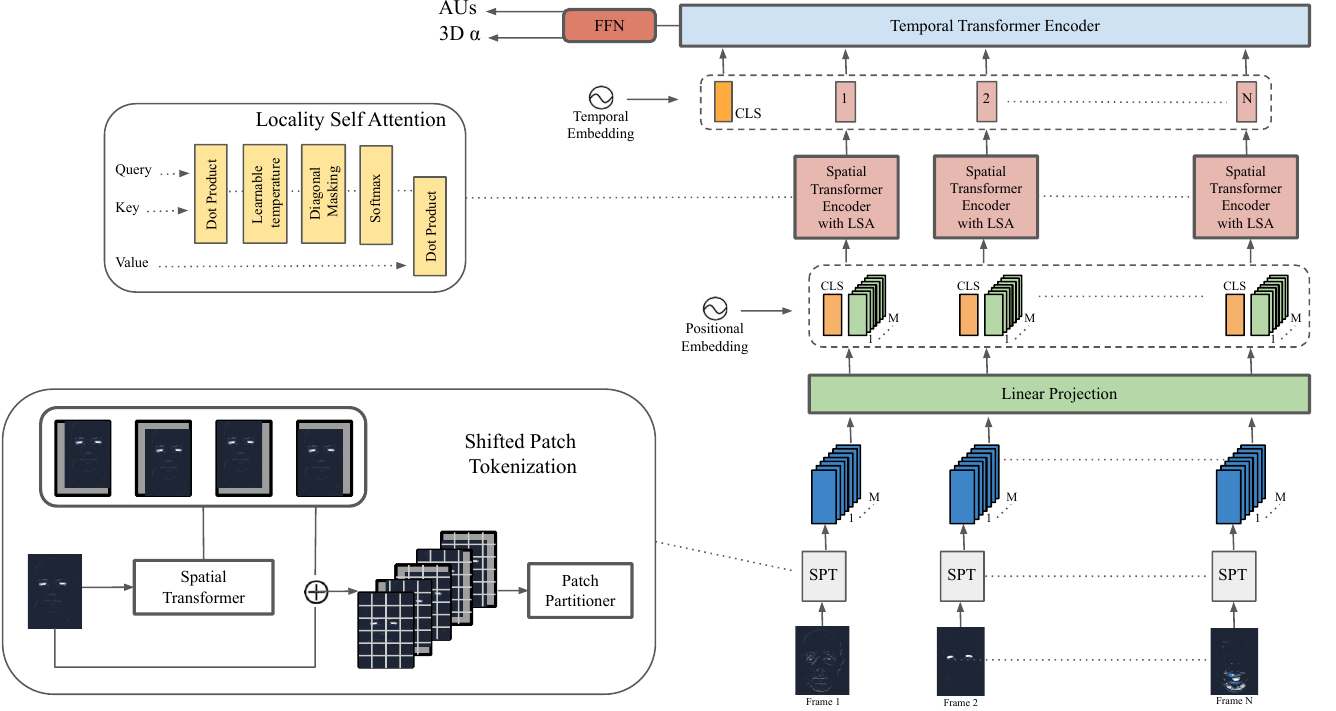}
		\caption{\revision{Architecture overview. Each video frame is first augmented using Shifted Patch Tokenization and divided into tokens. Tokens are linearly projected and fed to a spatial transformer with Locality Self Attention, along with a spatial CLS token. This operation is performed in parallel for each frame. We retain only the CLS token output for each frame and feed them to the temporal transformer, which captures temporal patterns and generates the final classification using a feed-forward network applied to the output of the temporal CLS token.}}
		\label{fig:architecture}
	\end{figure*}
	
	\section{Neuromorphic Action Unit Classification}
	\label{sec:problem_statement}
	To classify Action Units from event streams, we propose a multi-task model optimizing two losses. First, we minimize a classification loss $\mathcal{L}_{AU}$ for the main task at a video level, i.e. optimizing the probability of each class after having observed the whole sequence. The second loss $\mathcal{L}_{\alpha}$ is optimized for every frame in the video sequence, regressing the coefficients $\alpha_{AU}$ that define the face shape.
	More formally, our training setting is the following. Given a set of videos composed of $N$ event frames $F_t=F^1,..., F^N$, we supervise our model with a sequence of $N$ face shape coefficients $\alpha_{AU}^{*t}, t=1,..., N$ and a video class label $c^*_{AU}$, indicating the Action Unit performed in the video.
	The resulting loss is therefore $\mathcal{L} = \mathcal{L}_{AU} + \lambda \mathcal{L}_{\alpha}$, where $\lambda$ is a hyperparameter balancing the two losses.
	We use for classification a cross-entropy loss $\mathcal{L}_{AU} = -\sum_{i=1}^C c^*_{AU}log(c_i)$, and for regression a Mean Squared Error loss averaged over each frame $\mathcal{L}_{\alpha} = \frac{1}{N}\sum_{t=1}^N||\alpha_{AU}^t - \alpha_{AU}^{*t}||_2^2$, where $c_i$ is the logit for the $i-th$ Action Unit, C is the number of Action Units and $\alpha_{AU}^t$ is the 3D face shape coefficient vector predicted for frame $t$.
	
	\revision{
		We introduce our novel spatio-temporal vision transformer model that, taking inspiration from \citep{lee2021vision}, combines Shifted Patch Tokenization (SPT) and Locality Self-Attention (LSA) to improve spatial reasoning and, thanks to a multi-stage transformer, can integrate such information over time. Our approach aims to enhance the inductive locality bias and optimize attention mechanisms to better capture spatial and temporal information in event frames.
		
		The standard Vision Transformer (ViT) \citep{dosovitskiy2020image} tokenizes images by dividing them into non-overlapping patches and processing each patch independently. However, this approach can lead to a loss of important spatial information.
		To address this issue, \citep{lee2021vision} propose Shifted Patch Tokenization (SPT). In SPT, an input image is spatially shifted in four diagonal directions: left-up ($lu$), right-up ($ru$), left-down ($ld$), and right-down ($rd$). These shifts help capture a broader context around each pixel, increasing the receptive field of each token.
		After applying the shifts, the original image and its shifted versions are concatenated along the channel dimension. This augmented input is then flattened into a sequence of patch tokens, which are linearly projected into a higher-dimensional space. The SPT approach effectively embeds more spatial information into each token, enhancing the model’s ability to learn spatial relationships between neighboring pixels.
		
		More formally, let 
		$ x \in \mathcal{R}^{H\times W\times C}$
		denote an input image of height $H$, width $W$, and $C$ channels. We shift $x$ in the directions defined above, creating shifted versions $s_{lu}, s_{ru}, s_{ld}$ and $s_{rd}$.
		The shifted and original images are concatenated to form:
		\begin{equation}
			x_{shifts} = \left[ x ~|~ s_{lu} ~|~ s_{ru} ~|~ s_{ld} ~|~ s_{rd} \right] \in \mathcal{R}^{H\times W\times 5C}.
		\end{equation}
		This concatenated image is then divided into non-overlapping patches, which are linearly projected to generate the patch embeddings.
		
		To further improve the model’s performance, we use a Locality Self-Attention (LSA) mechanism, which modifies the standard self-attention in ViTs to emphasize local context. LSA is designed to reduce the over-smoothing of attention scores that can occur when the self-attention mechanism equally attends to all tokens, regardless of their spatial locality.
		LSA uses two key techniques: diagonal masking and learnable temperature scaling. Diagonal masking excludes self-tokens from the attention computation, preventing the model from focusing too heavily on individual tokens and encouraging it to attend to neighboring tokens instead. This is achieved by setting the diagonal elements of the attention score matrix to a large negative value. Learnable temperature scaling instead introduces a parameter that adjusts the softmax temperature dynamically, sharpening the attention distribution to make the model focus more on relevant tokens.
		Our model incorporates both SPT and LSA within a unified transformer architecture.
		
		The overall architecture consists of two main transformer components: a spatial transformer and a temporal transformer.
		The spatial transformer applies the LSA mechanism to process the spatial tokens generated by SPT, thereby capturing detailed spatial information effectively. Following the spatial processing, a transformer layer is used to integrate the temporal context, i.e., processing the outputs of the spatial transformer for each frame in the video sequence.
		To enable and efficient exchange of information between the temporal transformer (applied patch-wise to each frame) and the temporal transformer, we inject along the patch tokens also a CLS token, that is used to summarize the content of the frame. Therefore the temporal transformer only observes the sequence of spatial CLS tokens. An additional temporal CLS token is also used to obtain a final video-level semantic summary.
		In detail, each input image is first processed by SPT to generate enriched patch tokens. These tokens are then passed through the spatial transformer layers with LSA, where the model learns to focus on relevant local regions. The output of the spatial transformer for each patch is first pooled through class token pooling and then fed to the temporal transformer, which processes a sequence of frame-level features to account for temporal dependencies.
		We inject positional embeddings both in the spatial and the temporal transformer to inform the architecture of relative spatial and temporal positions of tokens.
		A final fully connected layer with layer normalization takes the transformed temporal CLS token and outputs the final classification.
		An overview of this architecture is shown in Fig. \ref{fig:architecture}.
	}

	\revision{In principle, to address the task of AU classification, any model capable of processing sequences of frames can be used. In order to compare our proposed model against other baselines we tested several alternatives, namely} (a) ResNet18+LSTM; (b) ResNet18+Transformer; (c) I3D.
	The first two architectures leverage a ResNet18 model, pre-trained on ImageNet, that acts as a backbone extracting 1024-dimensional features. We observed that, even if the model was trained on RGB data, the classification network still benefits from the pre-training. We freeze the convolutional part of the model, finetuning the fully connected layers and connecting them to either an LSTM or Transformer layer.
	For the LSTM model, we use three layers with hidden size of 256. The final hidden state is then fed for every timestep to a regression head composed of two fully connected layers with size 128 and 32 as the number of components describing the face shape to be regressed $\alpha_{AU}$. Similarly, a classification head with two fully connected layers with sizes 128 and 24, followed by a sofmax activation, generates a probability distribution over Action Units. This head is fed with the final hidden state of the LSTM, after the whole sequence has been processed.
	The transformer model operates in a similar way. The ResNet18 outputs for each frame are fed to a transformer encoder, yielding a sequence of outputs, which are then fed to regressors with the same structure as the ones in the LSTM model. Along with the input tokens of the transformer, we fed a CLS token, which, after being processed by the encoder, we use as input for a classification head.
	
	The I3D model instead, follows a different structure. We use a single branch Inception model with Inflated 3D convolutions \citep{carreira2017quo}. Here, the whole sequence of event frames is stacked together as a 3D tensor and processed to obtain a final 256-dimensional feature. As in the previous models we use an Action Unit classification head and a face shape regression head, however, since we do not process frames individually, we directly generate the concatenation of all the 3D coefficients, i.e. the final fully connected layer has an output dimension of $L \times 32$, where $L$ is the sequence length, that we fix to 75 frames (2.5 seconds).
	All fully connected layers except the final ones have ReLU activations in all the models. The models are trained with Adam using a learning rate of 0.001.

	\newcolumntype{H}{>{\setbox0=\hbox\bgroup}c<{\egroup}@{}}
	\begin{table}[t]
		\centering
		\resizebox{0.9\columnwidth}{!}{
			\begin{tabular}{@{}lHccc@{}}
				\toprule
				Model & $\alpha_{AU}$ representation & ~~~~~~Accuracy~~~~~~ & ~~~~~~top 3 Accuracy~~~~~~ & ~~~~~~top 5 Accuracy~~~~~~\\
				\midrule
				Transformer & Sequence & \textbf{69.27} & \textbf{82.76}	& \textbf{87.11}\\
				LSTM & Sequence & 50.34 & 82.36 & 86.66\\
				\bottomrule
			\end{tabular}
		}
		\caption{AU classification from 3D face deformation coefficients $\alpha_{AU}$.}
		\label{tab:AUalphas}
	\end{table}

	\section{Experiments}
	
	\subsection{Classification from Face Shapes}
	To assess the quality of our proposed dataset, Tab. \ref{tab:AUalphas}, shows the results of a control experiment without employing event data. 
	First, landmarks are extracted from the frames through face alignment \citep{bulat2017far}, and subsequently, the model described in Sec. \ref{subsec:3dmm_fitting} computes the deformation coefficients denoted as $\alpha_{AU}$. Finally, we feed the estimated coefficients to an Action Unit classifier.
	In Tab. \ref{tab:AUalphas}, models are trained to classify Action Units in videos, utilizing alpha coefficients for each frame.
	
	Treating the coefficients as a sequence, we trained both a Transformer (2 encoder layers; 2 decoder layers; 2 heads) and an LSTM model (1 layer with hidden size 256). Interestingly, the Transformer model achieved the best results ($69.27\%$ accuracy in Tab. \ref{tab:AUalphas}).
	\revision{This control experiment demonstrates that the collected videos carry a sufficiently informative signal to effectively estimate 3DMM coefficients with models pre-trained on datasets such as \citep{bulat2017far, Cosker2011AFV, VOCA2019}. Thanks to the expressiveness of 3DMM coefficients, simple models as the ones presented in Tab. \ref{tab:AUalphas} can be trained to classify Action Units effectively.
		Nonetheless, given the non-trivial nature of Action Unit classification, training models from scratch poses considerable challenges, as shown in the following sections.}

	
	\begin{table}[t]
		\centering
		\resizebox{0.9\columnwidth}{!}{
			\begin{tabular}{@{}lccccc@{}}
				\toprule
				Model~~~ & ~~~Mod~~~ & ~~~Accuracy~~~ & ~~~top 3 Accuracy~~~ & ~~~top 5 Accuracy~~~\\
				\midrule
				\multirow{2}{*}{ResNet18+LSTM} & Event& \textbf{46.23} & \textbf{62.91} & \textbf{68.58}\\
				&  RGB& 25.65 & 38.86 & 46.12 \\ \hline
				\multirow{2}{*}{ResNet18+Transf.} & Event&  \textbf{31.74} & \textbf{40.63} & \textbf{52.08}\\
				&RGB& 4.16 & 12.50 & 20.83\\ \hline
				\multirow{2}{*}{I3D} & Event& \textbf{47.08} & 69.58 & 80.66\\
				&RGB& 29.86 & \textbf{72.00} & \textbf{82.05}\\ \hline
				\multirow{2}{*}{Ours} & Event & \textbf{48.78} & \textbf{69.40} & \textbf{78.26} \\
				&RGB&  44.93  &  61.60  & 71.73\\ \hline
				\bottomrule
			\end{tabular}
		}
		\caption{\revision{Comparison of action unit classification accuracy for models trained with event data or RGB data. It emerges that events allow the models to learn more meaningful features for the task.}}
		\label{tab:eventVSrgb}
	\end{table}

	\begin{table*}[t]
		\centering
		\resizebox{1.0\textwidth}{!}{
			\begin{tabular}{@{}l|ccc|cccccccccccccccccccccccc@{}}
				\toprule
				\multicolumn{4}{c}{}  & \multicolumn{20}{c}{Accuracy (\%)}\\ 
				\midrule
				Model & ~~Acc~~ & ~~Top3~~ & ~~Top5~~ & ~~1~~~ & ~~2~~~ & ~~4~~~ & ~~6~~~ & ~~7~~~ & ~~9~~~ & ~~10~~ & ~~12~~ & ~~14~~ & ~~15~~ & ~~17~~ & ~~23~~ & ~~24~~ & ~~25~~ & ~~26~~ & ~~27~~ & ~~43~~ & ~~45~~ & ~~51~~ & ~~52~~ & ~~53~~ & ~~54~~ & ~~55~~ & ~~56~~ \\ \hline
				

				ResNet18+LSTM & 50.21 & 70.71 & \textbf{81.17} & \textbf{40.9} & 30.8 & 53.0 & 34.8 & 45.6 & 18.3 & 29.6 & 6.1 & 42.9 & \textbf{17.6} & 34.5 & 0.0 & \textbf{23.4} & 36.4 & 39.6 & 65.5 & \textbf{57.1} & \textbf{73.6} & 94.0 & 85.1 & 92.3 & \textbf{100} & \textbf{87.3} & 98.6 \\
				
				ResNet18+Transf. & 43.35   &   65.59 & 80.05 & 20.3 & 53.4 & 46.5 & 28.8 & 46.3 & \textbf{43.3} & 10.8 & 1.0 & 26.2 & 13.2 & 28.1 & 30.5 & 4.4 & \textbf{38.0} & 11.0 & 47.5 & 36.9 & 32.3 & 90.8 & 66.5 & \textbf{100} & 84.0 & 80.7 & \textbf{100} \\
				
				I3D & 49.58 & \textbf{71.08} & 79.41 & 29.6 & 45.1 & 40.1 & \textbf{40.0} & \textbf{50.7} & 13.4 & \textbf{62.7} & \textbf{60.2} & \textbf{43.3} & 1.7 & \textbf{41.4} & \textbf{36.0} & 4.2 & 37.2 & \textbf{65.6} & \textbf{70.9} & 39.1 & 62.6 & 57.3 & 84.4 & 63.9 & 90.2 & 75.8 & 77.1 \\
				
				Ours & \textbf{50.56} & 69.24 & 80.68 & 26.9 & \textbf{72.0} & \textbf{53.9} & 11.5 & 38.5 & 34.6 & 46.2 & 12.0 & 26.9 & 15.4 & 23.1 & 15.4 & 19.2 & 30.8 & 34.6 & 68.0 & 48.0 & 69.2 & \textbf{96.2} & \textbf{100} & \textbf{100} & 92.3 & 77.8 & \textbf{100}\\
				\bottomrule
				
			\end{tabular}
		}
		\caption{\revision{Accuracy for different models, averaged over the whole test set and reported per Action Unit.}}
		\label{tab:AU_acc}
	\end{table*}

	\newcommand{\meshwidth}{.101\linewidth}
	\begin{figure*}[t]
		\includegraphics[width=\meshwidth]{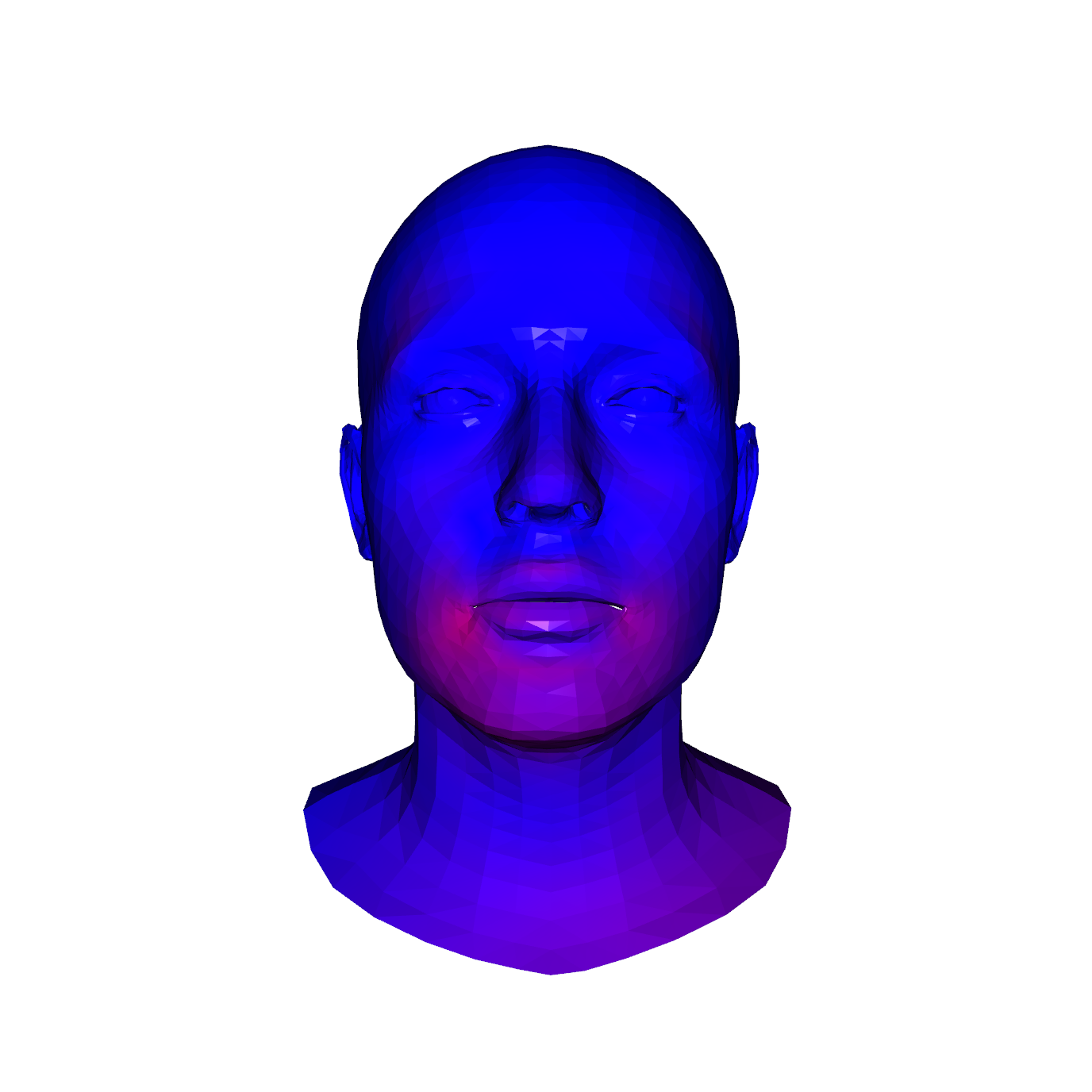}
		\includegraphics[width=\meshwidth]{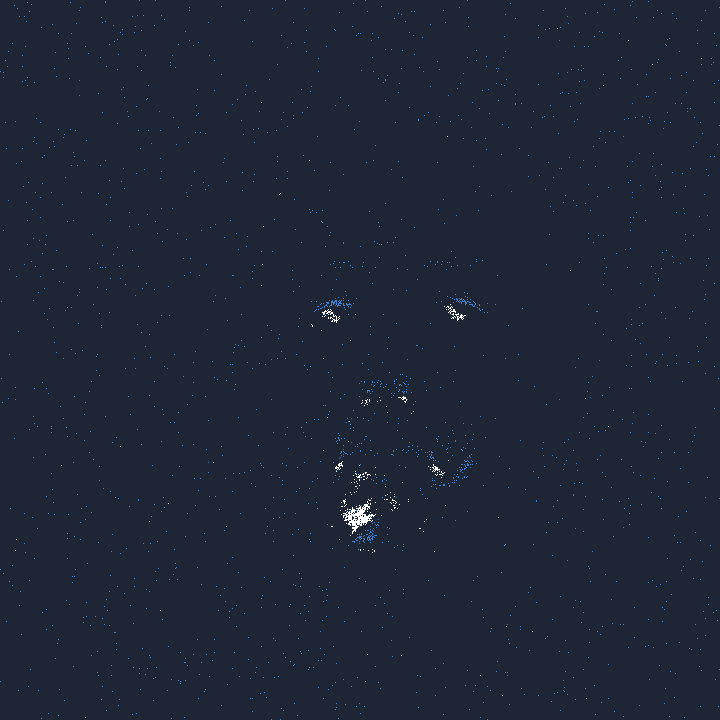} \includegraphics[width=\meshwidth]{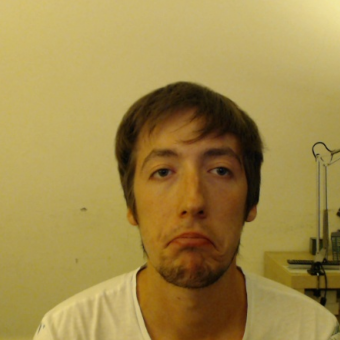}
		\includegraphics[width=\meshwidth]{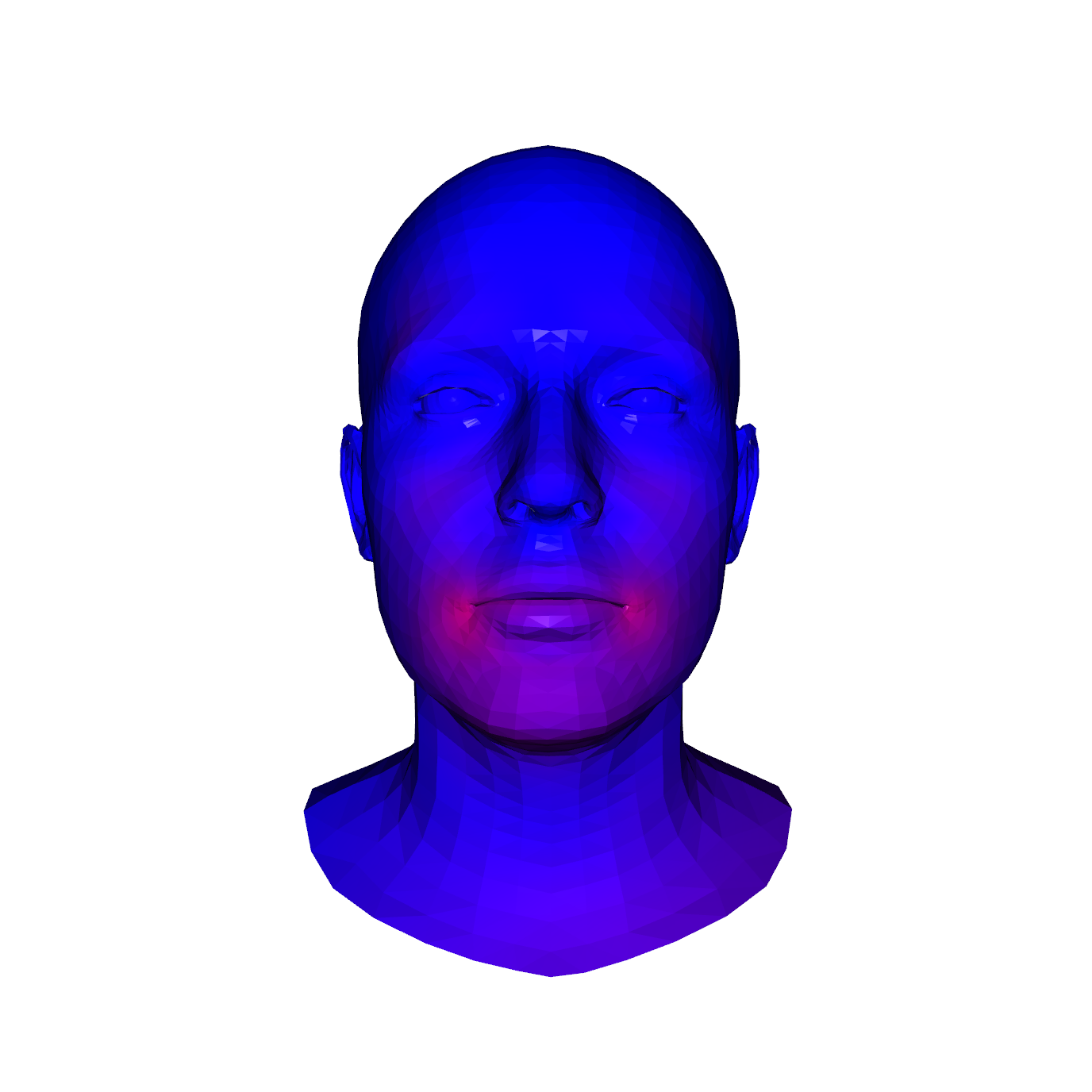}
		\includegraphics[width=\meshwidth]{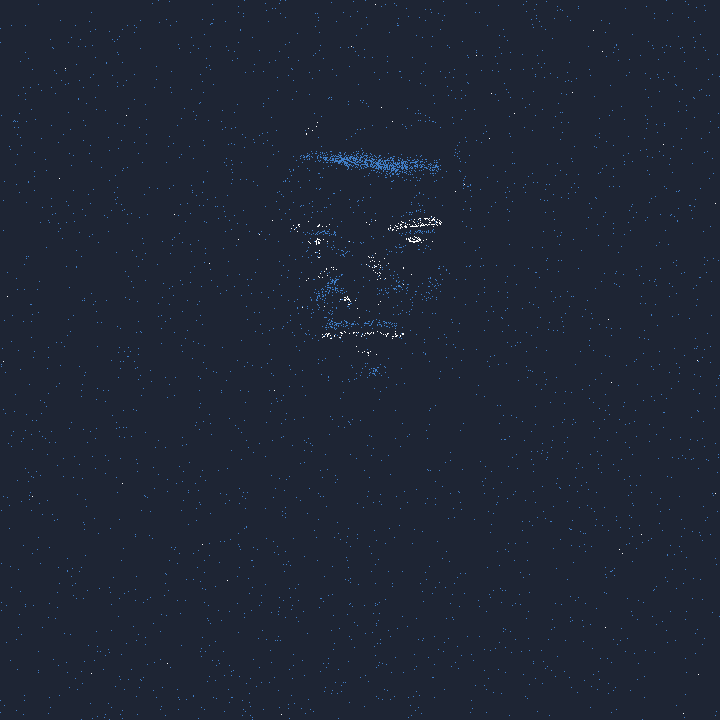} \includegraphics[width=\meshwidth]{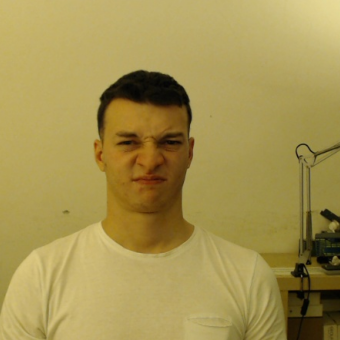}
		\includegraphics[width=\meshwidth]{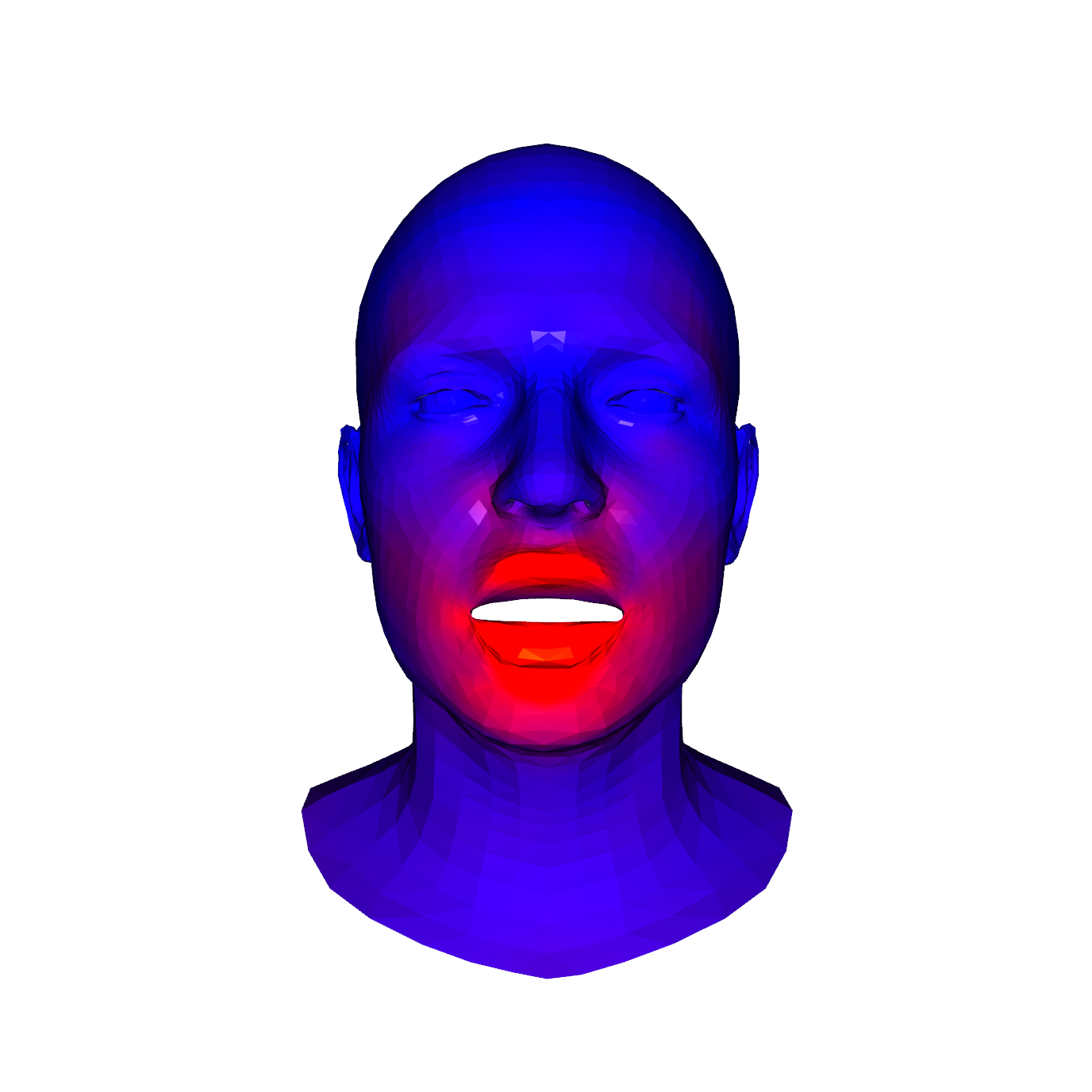}
		\includegraphics[width=\meshwidth]{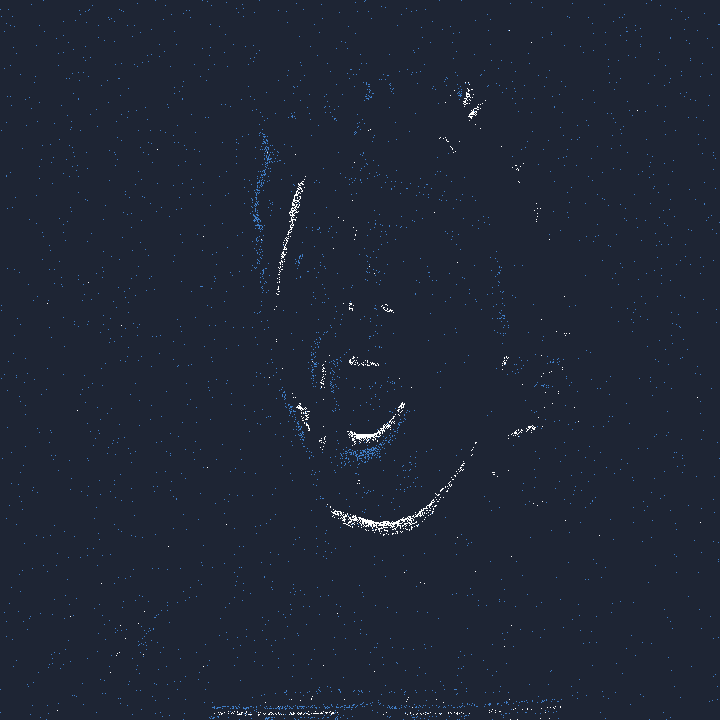} \includegraphics[width=\meshwidth]{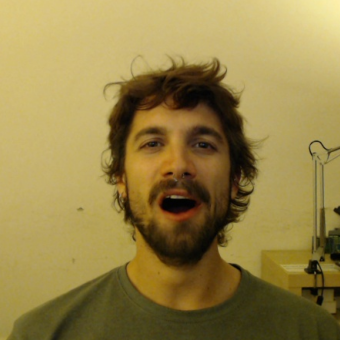}
		\caption{Samples of Action Units being performed and estimated 3D face shape. \textit{Left}: RGB frame; \textit{Center}: corresponding event frame; \textit{Right:} Reconstructed 3D mesh with the most active parts colored in red as distance from a neutral reference model.}
		\label{fig:qualitative_samples}
	\end{figure*}

	\subsection{Event vs RGB Comparison}
	To motivate the usage of neuromorphic data for action unit classification, we present in Tab. \ref{tab:eventVSrgb} a comparison between the models presented in Sec. \ref{sec:problem_statement}, trained with RGB or event data, without the usage of cross-supervision. It clearly emerges that the models trained with neuromorphic data outperform their RGB counterparts by a large margin. 
	\revision{Notably, our proposed model ("Ours" in table) with event data achieved the highest top-1 accuracy of 48.78\%, further validating the effectiveness of using neuromorphic data for this task.}
	The ResNet18+Transformer model struggles the most to address the task. We impute this to the fact that the transformer layer is trained from scratch and that it would likely require a larger amount of samples to be trained effectively. Surprisingly, the model when trained on RGB data does not learn effectively, yielding a top-1 accuracy which is equal to a random guess. This does not happen with event data, as the model reaches 31.74\%, which however is still lower than ResNet18+LSTM and I3D.
	\revision{Interestingly, our proposed spatio-temporal transformer model performs well even when trained with RGB data, reaching a top-1 accuracy of 44.93\%, which highlights its robustness across different modalities. However, a significant gap is still observable between the Event-based model and the RGB-based model.}
	\revision{A surprising fact is that the RGB variant of I3D, although not reaching good results at top-1, obtains the highest results for top-3 and top-5 accuracy. This finding suggests that 3D convolutional filters can provide a useful tool for analyzing subtle facial movements, allowing the model to at least recognize the AUs at least as macro-categories (e.g., AUs involving eye movements, head movements, etc.). However, the temporal granularity of the RGB frames does not allow the it to accurately discriminate between similar action units to obtain a good top-1 accuracy.}
	\revision{In conclusion, we believe that using event data helps training as events let the model focus on cues that are relevant to the task (the motion of facial parts) and the fine temporal granularity of the movements requires to be analyzed to accurately classify action units.}
	These findings confirm the importance of modeling facial dynamics with neuromorphic cameras rather than RGB data.

	\subsection{Results with Cross-Modal Supervision}
	In this section, we show the results obtained on the \dataset{} dataset.
	Tab. \ref{tab:AU_acc}  presents the Action Unit classification outcomes using event data for the multi-task models presented in Sec. \ref{sec:problem_statement}. For each model, the performance in terms of accuracy for all Action Units is also reported. 
	\revision{Notably, our proposed model outperforms the other baseline, achieving the highest Top-1 accuracy of 50.56\%, as well as strong Top-3 and Top-5 accuracies of 69.24\% and 80.68\%, respectively.} 
	\revision{Conversely, the model displaying the least favorable performance is ResNet18+Transformer, experiencing a drop of approximately 7\% compared to our proposed model} and ResNet18+LSTM and about 6\% compared to I3D. Despite this, the Top 5 accuracies of all the models are comparable, reaching an accuracy of approximately 80\%. In light of the inherently challenging nature of the task, all three models exhibit significant overall performance.
	
	\revision{Our model also demonstrates the ability to minimize common classification errors. Notably, Class 1 (Inner Brow Raiser) is frequently misclassified as Class 2 (Outer Brow Raiser), a common issue likely due to the subtle difference in brow movement. Additionally, Class 17 (Chin Raiser) is often mispredicted as Class 23 (Lip Tightener), suggesting overlapping muscle activity around the chin and lips. Furthermore, Class 12 (Lip Corner Puller) and Class 14 (Dimpler), as well as Class 14 (Dimpler) and Class 15 (Lip Corner Depressor), exhibit high misclassification rates, indicating difficulty in distinguishing between facial expressions involving lip and cheek movements. Finally, confusion between Class 23 (Lip Tightener) and Class 27 (Mouth Stretch) points to challenges in differentiating similar lip tightening and stretching actions.} 
	
	
	
	We also provide a qualitative analysis of the coefficients $\alpha_{AU}$,  generated by \revision{our model}. In Fig. \ref{fig:qualitative_samples} we show the RGB and event frame with the corresponding 3D face shape obtained by warping a neutral identity-free face model with the regressed $\alpha_{AU}$. We color-code the 3D mesh by highlighting the distance from the neutral reference face, hence showing the most active face parts.
	The ability of our approach to infer shape faces frame-by-frame thus provides a better characterization of the observed faces, as well as classifying the Action Units.

	Finally, we investigate the contribution of the cross-modal loss $\mathcal{L}_{\alpha}$ by training new models without it (Tab.~\ref{tab:ablation}). The additional supervision offered by the regression task over the coefficients $\alpha_{AU}$ shows consistent improvement for the AU classification.
	The positive impact of regressing the $\alpha_{AU}$ coefficients suggests that incorporating information from 3D face reconstruction helps the model better discern subtle nuances in facial movements associated with different AUs.

	\begin{table}[t]
		\centering
		\resizebox{0.9\columnwidth}{!}{
			\begin{tabular}{@{}lcccc@{}}
				\toprule
				Model~~~ &  ~~~$\mathcal{L}_{\alpha}$~~~ & ~~~Accuracy~~~ & ~~~top 3 Accuracy~~~ & ~~~top 5 Accuracy~~~\\
				\midrule
				ResNet18+LSTM & \cmark & 50.21 & 70.71 & \textbf{81.17}\\
				ResNet18+Transf. & \cmark & 43.35 & 65.59 & 80.05\\
				I3D & \cmark & 49.58 & \textbf{71.08} & 79.41\\
				Ours & \cmark & \textbf{50.56} & 69.24 & 80.68 \\
				\midrule
				ResNet18+LSTM & \xmark & 46.23 & 62.91 & 68.58\\
				ResNet18+Transf. & \xmark & 31.74 & 40.63 & 52.08\\
				I3D & \xmark & 47.08 & 69.58 & 80.66\\
				Ours & \xmark & 48.78 & 69.40 & 78.26 \\
				
				\bottomrule
			\end{tabular}
		}
		\caption{\revision{AU classification from Event images with and without the regression loss $\mathcal{L}_{\alpha}$.}}
		\label{tab:ablation}
	\end{table}
	
	\subsection{Emotion classification}
	\revision{To further validate our approach we apply our models to a downstream task such as emotion classification, by using the NEFER \citep{berlincioni2023neuromorphic} dataset. 
		The emotion classification data from NEFER presents two ground truth annotations per sample. Each sample records the reaction from a user while being shown particular videos. For each \textit{(user, video)} pair both the expected emotion (\texttt{A-priori}) and the one reported by the test subjects (\texttt{Reported}) are given.
		
		In order to extend our method to this task we replace the final classification linear layer, in our proposed model, with a newly initialized one shaped for the 8 recognized emotions and finetune the model.
		In Tab.~\ref{tab:nefer_accu} we compare the results of our best method against the baseline methods presented in \citep{berlincioni2023neuromorphic} and in \citep{berlincioni2024neuromorphic}. To better appreciate the benefits of pretraining on \dataset{}, we report results with and without it, showing that using a pretrained model can lead to a considerable improvement in terms of accuracy.}
			

	

	\begin{table}[t]
		\centering
		\resizebox{0.9\columnwidth}{!}{%
			\begin{tabular}{lcccc} 
				\hline
				Model & ~~~\makecell{FACEMORPHIC \\ Pretrain}~~~ & ~~~\makecell{Source}~~~ & ~~~\makecell{A-Priori \\ Labels}~~~   & ~~~\makecell{Reported\\ Labels}~~~  \\ 
				\toprule
				Conv3D~\citep{berlincioni2023neuromorphic}  & \xmark & RGB & 14.60 &  14.37  \\ 
				Conv3D~\citep{berlincioni2023neuromorphic} & \xmark & Event & 22.95 & 30.95 \\
				NVAE$_{f}$~\citep{berlincioni2024neuromorphic} & \xmark & Event & 19.20 & - \\
				NVAE$_{v}$~\citep{berlincioni2024neuromorphic} & \xmark & Event & 20.80 & -\\
				\midrule
				Ours  & \xmark & Event & 20.38 & 29.87 \\
				Ours  & \cmark & Event & \textbf{29.14} & \textbf{38.51} \\ \bottomrule
				
		\end{tabular}}
		\caption{Accuracy on NEFER \citep{berlincioni2023neuromorphic} of our proposed method, pretrained on \dataset{}. 
			\label{tab:nefer_accu}}
	\end{table}

	\section{\textbf{Inference Time}}
	\revision{Our proposed model is lightweight and can run in real-time. We observed an average inference speed of 18.8ms to process a 1-second long chunk, with events accumulated over a timespan of 33ms (i.e, we process 30 frames per second). Therefore, the model is capable of processing event streams in real-time.}
	
	\section{Conclusions and Future Works}
	We have presented the \dataset{} dataset, the first event-based Action Unit classification dataset, which is paired with temporally synchronized RGB footage. To perform Action Unit classification effectively, we leveraged a cross-modal supervision by extracting pose-invariant face shape coefficients from RGB frames using a 3D Morphable Model.
	In our experiments, we show that regressing such coefficients frame-per-frame, while training a video-level classifier, largely improves the overall classification accuracy. As a byproduct the model can also offer a better description of the face by producing a 3D reconstruction online, as the stream is processed.
	\revision{We also propose an innovative vision transformer architecture that integrates Shifted Patch Tokenization (SPT) and Locality Self-Attention (LSA) mechanisms with a spatio-temporal transformer, enabling a more effective capture of both spatial and temporal features from event-based data.
		The proposed model outperforms existing architectures, such as ResNet18+LSTM and I3D, in accurately classifying Action Units from neuromorphic data.}
	
	\revision{In addition, we demonstrate that \dataset{} can be used as an effective pre-training for downstream event-based tasks also on different datasets, as the results on emotion recognition on NEFER confirmed.}
	Future work directions should include the analysis of different encoding strategies for event streams, as they can heavily influence the data volume, its representation, and consequently the appropriate architecture.
	Similarly, a modeling of events with finer accumulation times could offer benefits in capturing less perceivable facial movements, at the cost of a higher computational burden.
	
	\section*{Acknowledgments}
	This work was partially supported by the European Commission under European Horizon 2020 Programme, grant number 951911—AI4Media. This work was partially supported by the Piano
	per lo Sviluppo della Ricerca (PSR 2023) of the University of Siena - project FEATHER: Forecasting and Estimation of Actions and Trajectories for Human-robot intERactions.

	\bibliographystyle{abbrvnat}
	\bibliography{egbib}  

		
		
		
		

\end{document}